# Fruit Classification System with Deep Learning and Neural Architecture Search

Christine Dewi[1,2], Dhananjay Thiruvady[1], and Nayyar Zaidi[1]

[1]School of Information Technology, Deakin University, Geelong, Australia, Email: c.dewi@deakin.edu.au, dhananjay.thiruvady@deakin.edu.au, nayyar.zaidi@deakin.edu.au,
[2]Department of Information Technology, Satya Wacana Christian University, 52-60 Diponegoro Rd, Salatiga City, 50711, Indonesia, Email: christine.dewi@uksw.edu
*Correspondence: christine.dewi@uksw.edu, c.dewi@deakin.edu.au

Abstract: The fruit identification process involves analyzing and categorizing different types of fruits based on their visual characteristics. This activity can be achieved using a range of methodologies, encompassing manual examination, conventional computer vision methodologies, and more sophisticated methodologies employing machine learning and deep learning. Our study identified a total of 15 distinct categories of fruit, consisting of class Avocado, Banana, Cherry, Apple Braeburn, Apple golden 1, Apricot, Grape, Kiwi, Mango, Orange, Papaya, Peach, Pineapple, Pomegranate and Strawberry. Neural Architecture Search (NAS) is a technological advancement employed within the realm of deep learning and artificial intelligence, to automate conceptualizing and refining neural network topologies. NAS aims to identify neural network structures that are highly suitable for tasks, such as the detection of fruits. Our suggested model with 99.98% mAP increased the detection performance of the preceding research study that used Fruit datasets. In addition, after the completion of the study, a comparative analysis was carried out to assess the findings in conjunction with those of another research that is connected to the topic. When compared to the findings of earlier studies, the detector that was proposed exhibited higher performance in terms of both its accuracy and its precision.

Keywords: Object Classification, CNN, Fruit Identification, YOLONAS, YOLOv8, Deep Learning.

## 1. Introduction

Fruit detection is a computer vision task that involves identifying and locating fruits within images or video frames. This task is a subset of object detection, which aims to identify and locate various objects in images or videos. Fruit detection has several practical applications, including agriculture [1], food quality control [2][3], and inventory management [4]. The classification of fruits is a significant challenge in supermarket settings due to the essentiality of categorizing fruits accurately for cashiers to ascertain their respective prices [5][6]. The issue at hand has been somewhat mitigated by implementing barcodes for packaged goods. Nonetheless, a significant portion of consumers still express a preference for personally selecting their desired products [7]. Certain types of fruits are not suitable for packaging with barcodes, necessitating the use of weighing methods instead. One potential approach could involve the implementation of unique codes for each fruit. However, the arduous task of memorizing barcodes may introduce the possibility of price inaccuracies [8]. An alternative approach could involve providing the cashier with an inventory booklet including visual representations and corresponding codes. Nevertheless, it is important to note that the process of flipping through the pages of the inventory booklet may result in a significant expenditure of time [9].

The process of classifying fruits entails grouping them into categories according to a variety of qualities. These criteria can include, but are not limited to, size, shape, color, flavor, and botanical features. This classification can serve several functions, such as those in agriculture, botany, nutrition, and the culinary arts. The use of deep learning for the detection of fruit offers several significant applications and benefits across a variety of fields. Models that utilize deep learning can be utilized to assist in the evaluation of the quality of fruits based on characteristics such as size, shape, color, and flaws. The use of this information can help farmers optimize the times and methods for harvesting their crops. The estimation of crop yields that can be obtained by fruit detection can be of use in the production planning and resource allocation processes. Tracking the inventory of fruits along the supply chain, reducing spoiling, and assuring on-time delivery to merchants are all possible with the help of fruit detection.

Deep learning has significantly advanced fruit detection [10], making it more accurate, efficient, and applicable in various industries, including agriculture [11], food processing [12], retail, and more. It allows for automation and scalability in tasks related to fruit counting, sorting, quality control, and yield estimation, ultimately leading to increased productivity and reduced costs. YOLOv8 is the most recent version in the YOLO series, demonstrating advancements that enhance the distinctive characteristics that led to the broad recognition of its previous iterations. Enhanced levels of precision and efficiency are attained by the adoption of a revolutionary architecture, underpinned by transformer models [13][14]. The YOLOv8 model is highly efficient at detecting objects because of its advanced training methodology, which involves the integration of knowledge distillation and pseudo-labeling techniques [15][16].

Neural Architecture Search (NAS) is a technique used in the field of deep learning and artificial intelligence that aims to mechanize the process of simulating neural network designs [17]. NAS was developed by researchers at the University of California, Santa Cruz. The objective of NAS is to discover neural network topologies that are effectively customized for a specific task or dataset, to do so while minimizing the requirement for labor-intensive human design and hyperparameter tweaking. Using NAS technology, the laborious process of constructing deep learning models may be streamlined and automated, and the technology can also be used to construct deep neural networks quickly and efficiently, with the ability to adapt these networks to fit specific production requirements [18][19].

The major contribution of this research article is to introduce an innovative fruit categorization system that utilizes computer vision techniques and to address the limitations to the greatest extent possible. Our study is pioneering research since we apply NAS to solve fruit classification difficulties and our work improves upon a previous research study by 7% from 93.78% to 99.98%. Furthermore, it is anticipated that the suggested classifier will possess the capability to identify a wide range of fruit varieties. This study encompasses a total of 15 different varieties of fruits. Moreover, we not only extract traditional color and shape features but also incorporate crucial texture features. The anticipated classifier is projected to exhibit high accuracy through the utilization of a YOLO with Neural Architecture Search (NAS). This is due to the YOLO efficacy as a supervised classifier, which enables it to effectively categorize patterns that are nonlinearly separable and approximate any continuous function. Our work performs a comparative analysis of the YOLONAS with other YOLO models such as YOLOv5, YOLOv7, and YOLOv8.

The subsequent part provides a description and evaluation of relevant literature that precedes the current study. The third section presents a comprehensive review of the methods that we have proposed. Section 4 provides an overview of the dataset utilized in the study, as well as the training data employed for the system. Additionally, it presents the results obtained during the system's testing phase. Section 5 closes by providing recommendations for further research and development.

## 2. Materials and Methods
### 2.1. Fruit Detection with Deep Learning

Fruit detection with deep learning involves using neural network-based models to automatically identify and locate fruits within images or videos. In their study, Hong et al. [20] utilized morphological analysis as a means of categorizing walnuts and hazelnuts into three distinct groupings. The initial use of data fusion for nondestructive imaging of fresh intact tomatoes was conducted by Baltazar et al. [21], who subsequently employed a three-class Bayesian classifier. In their study, Pennington et al. [22] employed a clustering technique to classify fruits and vegetables. In their study, Pholpho et al. [23] employed visible spectroscopy to differentiate between non-bruised and bruised longan fruits. They further enhanced their analysis by incorporating principal component analysis (PCA), Partial Least Square Discriminant Analysis (PLS-DA), and Soft Independent Modeling of Class Analogy (SIMCA) techniques to construct classification models.

Fruit detection with deep learning can automate tasks such as fruit counting, sorting, quality assessment, and yield estimation, leading to increased efficiency and accuracy in various industries. The choice of architecture and the quality of the dataset are critical factors in achieving accurate fruit detection results. Deep learning models utilized in object detection want significant computational resources during the training phase. Therefore, it is advisable to contemplate the utilization of cloud-based services or specialized hardware, such as Graphics Processing Units (GPUs) or Tensor Processing Units (TPUs), to ensure efficient training. It is imperative to acknowledge that although deep learning models have demonstrated significant efficacy in fruit detection, they possess inherent limits, particularly in demanding circumstances characterized by significantly occluded or damaged fruits. One potential avenue for enhancing the robustness of fruit identification systems is through the integration of deep learning with other computer vision techniques or the incorporation of supplementary sensors [24][25]. Table 1 describes the benefits of fruit detection in various industries.

Table 1. Benefits of fruit detection in various industries.

| Various Industries | Benefits |
| --- | --- |
| Agriculture | (1) Crop Management: Fruit detection can be used for monitoring and managing fruit crops. Farmers can assess crop health, estimate yields, and plan harvesting schedules more effectively [26]. (2) Fruit Counting: Automated fruit counting can help in estimating crop yield, allowing for better resource allocation and harvest planning [27]. (3) Ripeness Assessment: Fruit detection can determine the ripeness of fruits, enabling selective harvesting to maximize quality [28]. |
| Food Processing and Quality Control | (1) Quality Assessment: Fruit detection systems can inspect and assess the quality of fruits based on various factors such as size, shape, color, and blemishes [29]. (2) Sorting and Grading: Automated sorting and grading systems can categorize fruits based on quality criteria, ensuring consistent product quality. (3) Reduced Waste: By identifying defective or damaged fruits early in the production process, food waste can be minimized. |
| Inventory Management | (1) Stock Control: Fruit detection can be used in warehouses and distribution centers to manage inventory levels and monitor the condition of stored fruits [30]. (2) Supply Chain Optimization: Accurate fruit detection can optimize the supply chain by ensuring that the right quantity and quality of fruits are available when and where needed [24]. |
| Retail and Consumer Applications | (1) Consumer Experience: In retail settings, fruit detection can enhance the consumer experience by providing information about the freshness, origin, and nutritional content of fruits [31]. (2) Checkout Automation: Automated checkout systems can use fruit detection to identify and price fruits accurately without manual input. |
| Research and Analysis | (1) Data Collection: Fruit detection technology can assist researchers in collecting data on fruit populations in natural ecosystems for ecological studies [32]. (2) Scientific Research: Fruit detection can support scientific research in fields such as botany and ecology by automating the identification of fruits in images and videos [33]. |
| Environmental Monitoring | Wildlife Conservation: Fruit detection can be used to monitor fruit availability for wildlife, aiding in conservation efforts by understanding animal food sources and migration patterns [34]. |
| Efficiency and Cost Savings: | Labor Savings: Automating fruit detection can reduce the need for manual labor in tasks like fruit counting, sorting, and quality control. |
| | Resource Optimization: Precise fruit detection allows for efficient resource allocation, such as water and fertilizers in agriculture [35]. |
| Health and Nutrition | Dietary Tracking: Fruit detection technology can assist individuals in tracking their fruit consumption and nutritional intake for better health management [36]. |

Fruit detection has the potential to increase efficiency, reduce waste, and improve the quality of fruit-related processes in agriculture, food processing, retail, and beyond. It also has the advantage of providing valuable data that can be used for decision-making and research purposes.

## 2.2. YOLONAS architecture

The fundamental architecture of YOLOv8 closely resembles that of YOLOv5, except for the substitution of the C3 module with the C2f module. This module is based on the concept of Constraint Satisfaction Problems (CSP). The C2f module of YOLOv8 was developed by integrating the ELAN concept from YOLOv7 with the C3 module. The purpose of this activity was to facilitate the development of the module. The integration was conducted to enhance the gradient flow information of YOLOv8 while ensuring the preservation of its lightweight design without compromising any aspect [37]. The SPPF module, which had a position of dominance, was employed exclusively during the final phase of the backbone design. Subsequently, a consecutive implementation of three wax pools, each measuring 5 by 5 inches, was performed. Subsequently, the outputs of each layer were merged to achieve precise identification of objects at different scales, while also maintaining a streamlined and efficient structure. The objective was achieved while maintaining a high level of precision [38].

NAS is an automated procedure that is used to design the architecture of neural networks to achieve the highest possible level of performance for a specific task. The purpose is to design an architectural framework that makes the least amount of use of human interaction and uses only the available resources. The overall structure of the NAS is depicted in Figure 1.

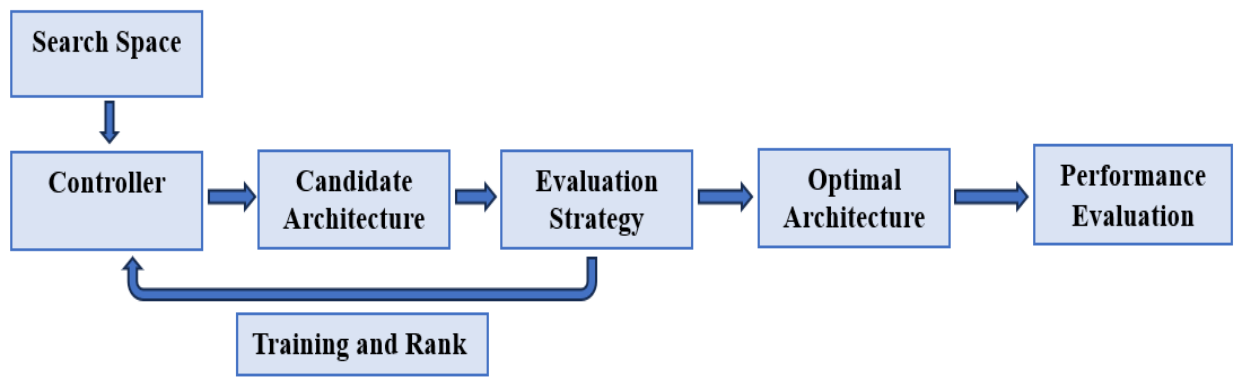

Figure 1. The architecture of NAS general framework.

The landscape of NAS algorithms is generally characterized by a considerable degree of complexity and lack of clarity [39]. According to the prevalent classification approach, NAS systems can be broken down into the following three basic components: (1) The area that is now being looked over. (2) The search process includes both the selection of the controller type as well as the evaluation of probable candidates [40]. (3) The method of judging an individual's performance. The phrase "search strategy" refers to the methodical process that is utilized to identify the architecture that is the most appropriate for the space that is being searched. Random search, reinforcement learning, evolutionary algorithms, sequential model-based optimization, and gradient optimization are the five key categories that may be used to classify NAS algorithms depending on the search technique that they take [41][42].

Deci AI [43]has made a substantial contribution to this subject by developing the object detection fundamental model known as YOLO-NAS, which represents a significant improvement in this sector. The product is the result of employing powerful Neural Architecture Search technology. This technology has been meticulously developed to particularly target and overcome the limits seen in previous rounds of YOLO models. With noteworthy improvements in quantization support as well as accuracy-latency trade-offs, YOLO-NAS exhibits a significant step forward in object detection [44][45]. To attain the highest possible level of performance, the YOLO-NAS architecture makes use of quantization-aware blocks as well as strategies for selective quantization. There is almost no reduction in precision when the model is converted to its INT8 quantized form, which represents a significant improvement in comparison to other models. The aforementioned developments lead to the creation of an improved architectural framework that demonstrates remarkable performance and capabilities that are unrivaled in object recognition [46][47].

To finish the labeling process, our work applies a total of three different labels, each of which is assigned a number between 0 and 2. YOLO's input values do not employ object coordinates to convey the data, in contrast to the input formats used by other models. This is because YOLO does not support them. The coordinates of the object's center point are included in the Yolo input data alongside the dimensions of the object's width and height (x, y, w, h). Common methods of representing bounding boxes include the use of either two coordinates, (x1, y1) and (x2, y2), or a single coordinate, (x1, y1), in conjunction with the width (w) and height (h) of the bounding box. The process of transformation is depicted using the Equations (1)-(6).

$$dw = 1/W \tag{1}$$

$$x = \frac{x1+x2}{2} \times dw \tag{2}$$

$$dh = 1/H \tag{3}$$

$$y = \frac{y1+y2}{2} \times dh \tag{4}$$

$$w = (x2 - x1) \times dw \tag{5}$$

$$h = (y2 - y1) \times dh \tag{6}$$

In this equation, $W$ represents the width of the image, while H represents its height. A corresponding text file with the same name as each image file in the same directory will be generated as a.txt file. Each a.txt file includes the object class, object coordinates, the height and width of the associated picture file, as well as additional metadata. Convolution, batch normalization, and SiLu activation functions for the YOLOv8 architecture are the three basic components that make up the convolutional neural network (CNN).

The incorporation of Attention Mechanism, Knowledge Distillation, and Distribution Focal Loss into the YOLO-NAS training process results in an improvement to the efficacy of such method [48]. The program demonstrates full interoperability with sophisticated inference engines like NVIDIA TensorRT and provides support for INT8 quantization. As a result, it achieves unrivaled runtime performance. YOLO-NAS shows remarkable performance in real-world scenarios, such as autonomous vehicles, robotics, and video analytics applications, where the ability to decrease latency and improve processing is of the utmost importance. However, this list is not exhaustive. The architecture of YOLO-NAS is depicted in Figure 2 [49][50].

YOLO-NAS typically follows a neural architecture search pipeline, which involves searching for the best neural network architecture to maximize both accuracy and efficiency. The architecture search process may include the following steps: (1) Search Space Definition: Define a search space of neural network architectures, including choices for various architectural elements such as the number of layers, layer types, kernel sizes, and other hyperparameters [51]. This search space can be quite extensive [52]. (2) Objective Function: Define an objective function or metric to evaluate the performance of candidate architectures. In the context of YOLO, this might include metrics like mean average precision (mAP), latency, and model size [53]. (3) Architecture Search: Employ a search algorithm to explore the defined search space and find the best-performing architectures. This search can be performed through techniques like reinforcement learning, evolutionary algorithms, or gradient-based optimization. (4) Evaluation: For each candidate architecture, evaluate its performance on a validation dataset to calculate the defined objective function. (5) Pruning: Prune fewer promising architectures to focus on more promising ones, optimizing the search process. (6) Fine-Tuning: After selecting the best architecture, fine-tuning to further improve its performance [54]. (7) YOLO Integration: Incorporate the optimized architecture into the YOLO framework for object detection. YOLO-NAS aims to strike a balance between accuracy and efficiency, ensuring that the resulting architecture is suitable for real-time object detection tasks, as YOLO is known for its speed and efficiency [55][56].

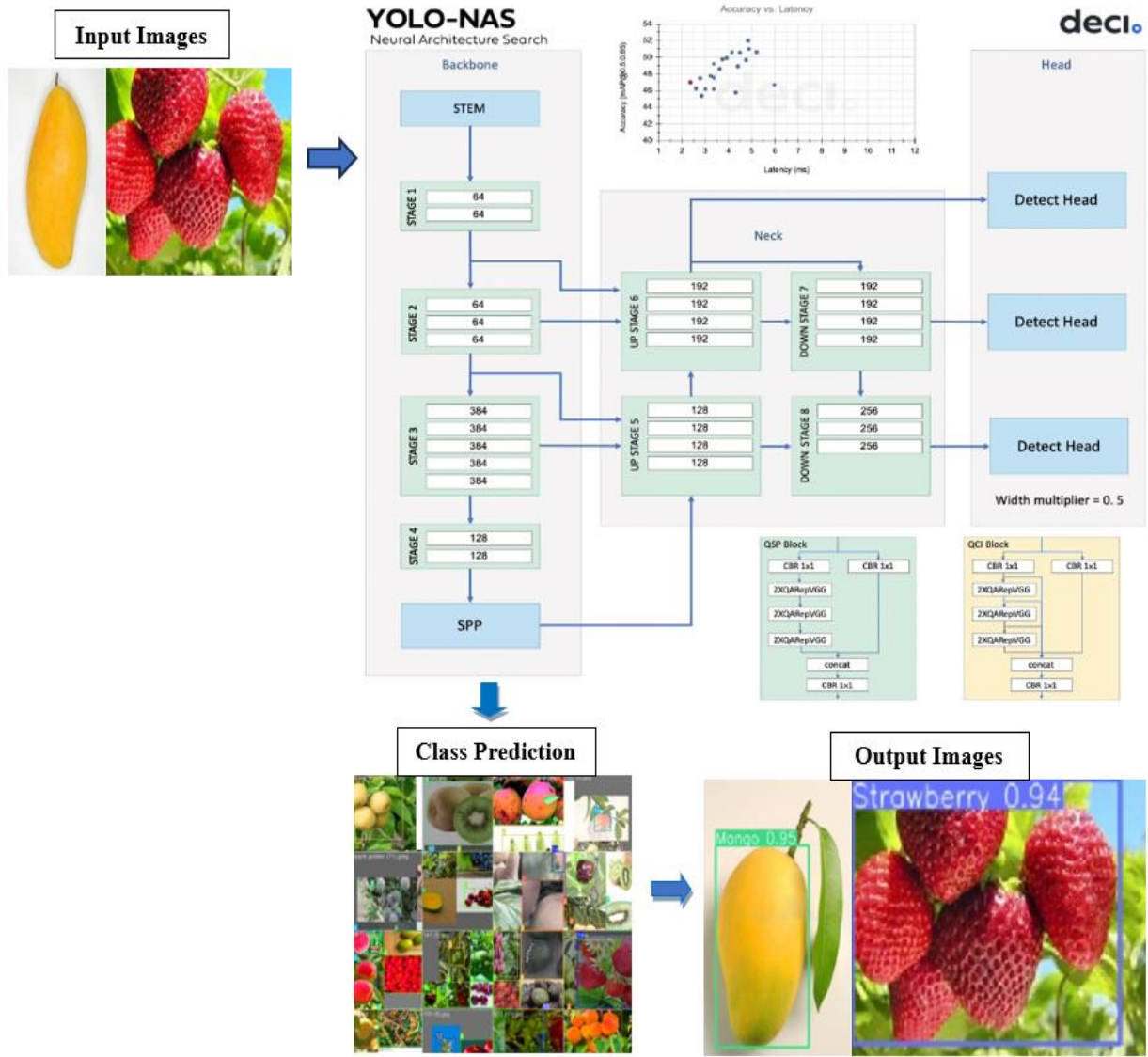

Figure 2. System Architecture of YOLONAS including the backbone, neck, and head.

YOLO-NAS consists of a backbone, neck, and head, which closely resemble several other models. The backbone algorithm derives distinctive characteristics from the input image. YOLO-NAS utilizes a dual-path backbone that consists of both dense feature extraction and sparse feature fusion paths. The dense path consists of several convolutional layers that incorporate skip connections within dense blocks. On the other hand, the sparse path includes transition blocks that decrease the spatial resolution and increase the number of channels. Both routes are interconnected with cross-stage partial connections, allowing for the movement of gradients. The neck layer amplifies the characteristics retrieved by the backbone, generating predictions at different scales. The YOLO-NAS system employs a multi-scale feature pyramid and backbone features to generate predictions at many scales, including small, medium, and large. The system's neck employs up-sampling and concatenation to merge data from different levels, hence expanding the receptive field. The head is responsible for carrying out the last stages of classification and regression tasks. The YOLO-NAS head is composed of two branches that run parallel to each other. One branch is dedicated to classification, while the other branch focuses on regression and utilizes generalized IoU loss. The classification branch determines the likelihood of each anchor box belonging to a specific class, while the regression branch calculates the precise coordinates of the bounding box.

## 3. Experimental Evaluation
### 3.1. Fruit Dataset

In collecting data for the detection of 15 types of fruit taken from Kaggle [57] and Roboflow [58]. The dataset of fruit images has a total of 2909 photographs, encompassing various fruit varieties such as Avocado, Banana, Cherry, Grape, Apple Braeburn, Apple golden 1, Apricot, Kiwi, Mango, Orange, Papaya, Peach, Pineapple, Pomegranate, and Strawberry. The Fruit Dataset sample is depicted in Figure 3.

The statistical data pertaining to the fruit category is presented in Table 2. Moreover, we divided our dataset into 70% of training, 20% of validation, and 10% of testing during our experiments.

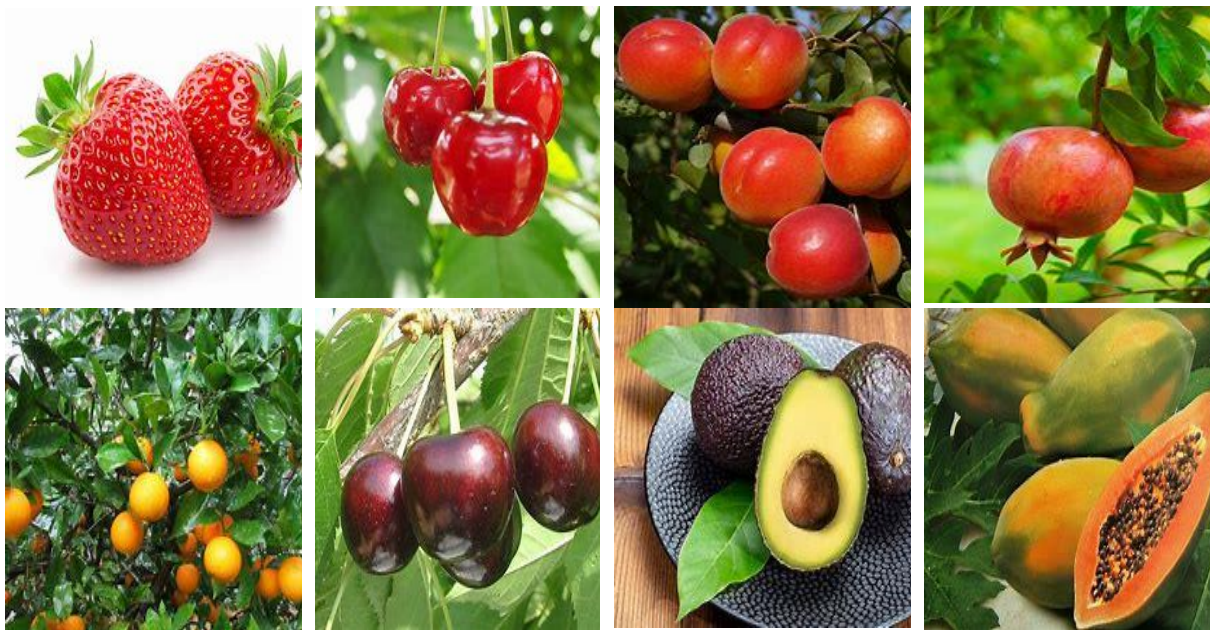

Figure 3. Fruit Dataset Example.

Table 2. Fruit Dataset Statistics.

| Class | Train | Valid | Test | Total |
|---|---|---|---|---|
| Apple Braeburn | 146 | 37 | 17 | 200 |
| Apple golden 1 | 116 | 37 | 22 | 175 |
| Apricot | 88 | 39 | 10 | 137 |
| Avocado | 152 | 39 | 18 | 209 |
| Banana | 141 | 47 | 19 | 207 |
| Cherry | 141 | 44 | 22 | 207 |
| Grape | 135 | 39 | 11 | 185 |
| Kiwi | 124 | 38 | 18 | 180 |
| Mango | 177 | 65 | 35 | 277 |
| Orange | 141 | 36 | 25 | 202 |
| Papaya | 122 | 38 | 19 | 179 |
| Peach | 149 | 32 | 19 | 200 |
| Pineapple | 148 | 40 | 21 | 209 |
| Pomegranate | 121 | 28 | 19 | 168 |
| Strawberry | 125 | 31 | 18 | 174 |

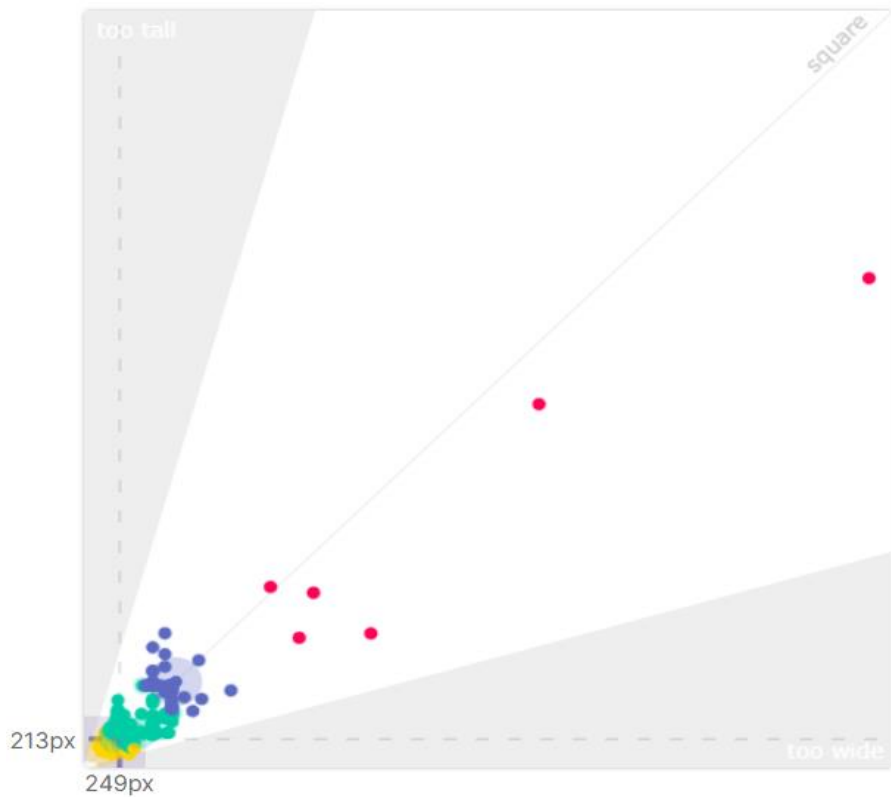

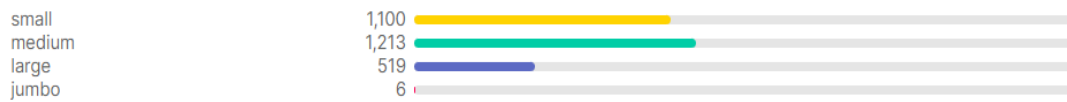

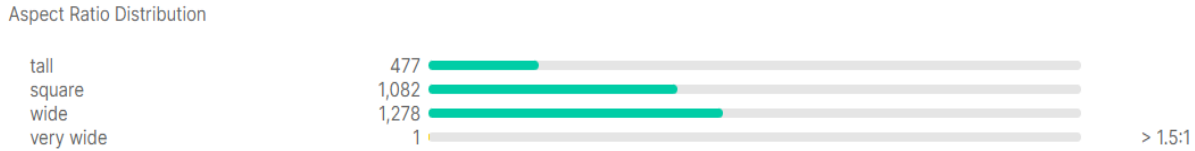

Figure 4. Dimension Insight of Fruit Dataset.

Figure 4 illustrates the dimensional aspects of the Fruit Dataset. The dataset has images of varying sizes, ranging from around 213 pixels to 249 pixels. It consists of 1100 instances of small photos, 1213 instances of medium images, 519 instances of large images, and 6 examples of jumbo images. The distribution of aspect ratios in our dataset reveals that there are 477 tall photographs, 1082 square images, 1287 broad images, and 1 very wide image.

### 3.3. Training Result

Data augmentation is a strategy that is widely used in the fields of machine learning and deep learning. The goal of this method is to artificially increase the variety of a training dataset by making a few modifications to the data that was initially collected. This can be performed through a process that is referred to as "data augmentation." With data augmentation, the generalization and robustness of a machine learning model can be increased by providing the model with a larger variety of training data. This is accomplished by supplying the model with more data. During the process of training, we will supplement the data by utilizing several different methods, including padding, cropping, and horizontal flipping, amongst others. Because of the many benefits that these methods offer, they are frequently used in the building of massive neural networks.

Moreover, the training model environment consists of an Nvidia RTX3060Ti GPU accelerator equipped with 11 gigabytes of RAM, an i7 Central Processing Unit (CPU), and 16 gigabytes of DDR2 memory. The training method for the YOLONAS system is conducted using a solitary graphics processing unit (GPU), with a key objective being the attainment of real-time detection capabilities. While the remaining 30% of the data is allocated for testing purposes, the remaining 70% is employed for training objectives. The YOLO method has been developed to predict multiple bounding boxes within every grid cell. In the training phase, it is preferable to allocate the task of predicting the bounding box for each item to a single predictor. The YOLO algorithm assigns a singular predictor as the "responsible" entity for object prediction, based on the selection of the prediction with the highest Intersection over Union (IOU) value compared to the ground truth. This tendency leads to a specialization among the predictors of bounding boxes. The improvement in the overall recall score is attributed to the predictive capabilities of each predictor in accurately anticipating items of specified sizes, aspect ratios, or categories.

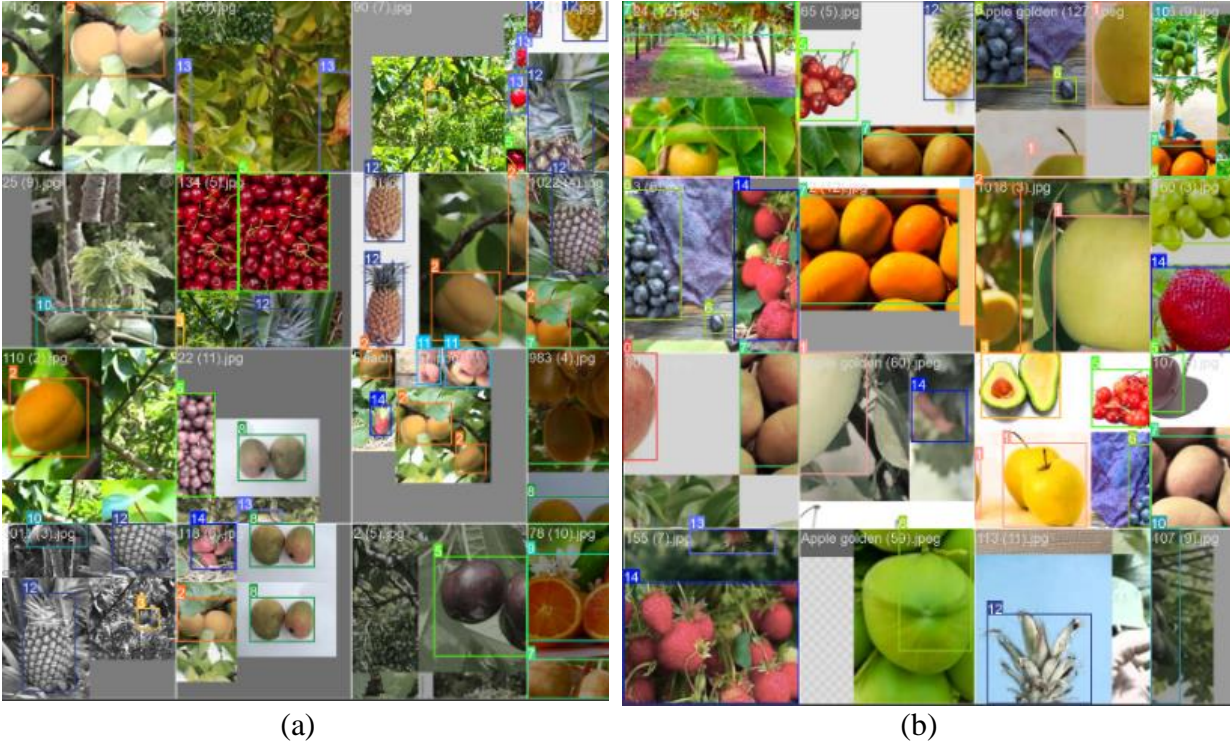

(a)                                                       (b)

Figure 5. Training process with YOLONAS_l (a) Batch 0 and (b) Batch 1.

Figure 5 presents a graphical depiction of the sequential stages encompassed within the training procedure for the labels of test batch 0 and the predictions of test batch 1 for YOLONAS_l. The utilization of NMS (Non-Maximum Suppression) has been found to enhance the accuracy of object detection after the processing stage. In the context of object detection, it is common practice to construct multiple bounding boxes for an image item. While there may be overlap or variation in the positioning of these bounding boxes, it is important to note that they all pertain to the identical item.

The key features of YOLONAS can be outlined as follows: The YOLO-NAS paper presents a novel basic block that is designed to be compatible with quantization techniques. This addresses a notable drawback observed in earlier YOLO models. The YOLO-NAS model utilizes sophisticated training techniques and post-training quantization methods to improve its overall performance. The YOLO-NAS model employs AutoNAC optimization and undergoes pre-training on well-known datasets including COCO, Objects365, and Roboflow 100. The utilization of pre-training renders it very appropriate for object detection jobs in production situations.

Table 3 exhibits the fruit dataset performance for YOLOv8n, YOLOv8s, and YOLOv8m. YOLOv8m achieves the highest mAP during training with 99.1% followed by YOLOv8s with 98.9% mAP and YOLOv8n with 97.8% mAP. Table 4 shows the experiment result with YOLOv5 and YOLOv7. YOLOv5n achieves better mAP 86% than YOLOv7that only got 56% mAP.

Table 3. Fruit dataset training performance for YOLOv8

| Class | YOLOv8n | | | YOLOv8s | | | YOLOv8m | | |
|---|---|---|---|---|---|---|---|---|---|
| | P | R | mAP | P | R | mAP | P | R | mAP |
| All | 0.961 | 0.949 | 0.978 | 0.977 | 0.971 | 0.989 | 0.974 | 0.975 | 0.991 |
| Apple B | 0.953 | 0.939 | 0.967 | 0.993 | 0.971 | 0.993 | 0.988 | 0.955 | 0.991 |
| Apple GI | 0.986 | 0.953 | 0.975 | 0.99 | 0.964 | 0.992 | 0.986 | 0.971 | 0.992 |
| Apricot | 0.956 | 0.974 | 0.988 | 0.974 | 0.985 | 0.994 | 0.971 | 0.998 | 0.994 |
| Avocado | 0.969 | 0.98 | 0.993 | 0.972 | 0.996 | 0.994 | 0.973 | 0.99 | 0.994 |
| Banana | 0.973 | 0.981 | 0.981 | 0.978 | 0.976 | 0.975 | 0.975 | 0.976 | 0.981 |
| Cherry | 0.98 | 0.981 | 0.985 | 0.981 | 0.989 | 0.985 | 0.981 | 0.989 | 0.982 |
| Grape | 0.986 | 0.974 | 0.982 | 0.952 | 0.981 | 0.974 | 0.977 | 0.981 | 0.985 |
| Kiwi | 0.925 | 0.947 | 0.975 | 0.944 | 0.982 | 0.99 | 0.938 | 0.986 | 0.99 |
| Orange | 0.971 | 0.9 | 0.972 | 0.976 | 0.945 | 0.99 | 0.973 | 0.946 | 0.991 |
| Mango | 0.899 | 0.858 | 0.937 | 0.979 | 0.902 | 0.984 | 0.947 | 0.928 | 0.985 |
| Papaya | 0.937 | 0.937 | 0.977 | 0.971 | 0.963 | 0.993 | 0.966 | 0.966 | 0.992 |
| Peach | 0.988 | 1 | 0.994 | 0.99 | 1 | 0.995 | 0.989 | 1 | 0.995 |
| Pineaple | 0.977 | 0.953 | 0.99 | 1 | 0.976 | 0.995 | 0.99 | 0.977 | 0.994 |
| Pomegranate | 0.959 | 0.908 | 0.971 | 0.972 | 0.978 | 0.993 | 0.973 | 0.98 | 0.993 |
| Strawberry | 0.955 | 0.968 | 0.989 | 0.975 | 0.974 | 0.994 | 0.984 | 0.974 | 0.994 |

Table 4 Fruit dataset training performance for YOLOv5 and YOLOv7

| Class | YOLOv5n | | | YOLOv7 | | | YOLOv7x | | |
|---|---|---|---|---|---|---|---|---|---|
| | P | R | mAP | P | R | mAP | P | R | mAP |
| All | 0.827 | 0.8 | 0.86 | 0.542 | 0.556 | 0.566 | 0.552 | 0.523 | 0.549 |
| Apple B | 0.805 | 0.767 | 0.853 | 0.462 | 0.46 | 0.478 | 0.566 | 0.364 | 0.497 |
| Apple GI | 0.905 | 0.865 | 0.955 | 0.614 | 0.703 | 0.702 | 0.65 | 0.757 | 0.761 |
| Apricot | 0.725 | 0.5 | 0.674 | 0.499 | 0.35 | 0.401 | 0.464 | 0.519 | 0.47 |
| Avocado | 0.787 | 0.841 | 0.839 | 0.644 | 0.386 | 0.485 | 0.658 | 0.318 | 0.441 |
| Banana | 0.952 | 0.923 | 0.956 | 0.756 | 0.476 | 0.689 | 0.729 | 0.346 | 0.55 |
| Cherry | 0.855 | 0.754 | 0.838 | 0.56 | 0.649 | 0.648 | 0.535 | 0.491 | 0.542 |
| Grape | 0.84 | 0.91 | 0.933 | 0.619 | 0.696 | 0.748 | 0.575 | 0.56 | 0.673 |
| Kiwi | 0.897 | 0.778 | 0.89 | 0.695 | 0.578 | 0.656 | 0.734 | 0.578 | 0.669 |
| Orange | 0.67 | 0.727 | 0.687 | 0.43 | 0.409 | 0.341 | 0.455 | 0.295 | 0.274 |
| Mango | 0.702 | 0.625 | 0.717 | 0.55 | 0.732 | 0.696 | 0.452 | 0.714 | 0.601 |
| Papaya | 0.926 | 0.941 | 0.96 | 0.447 | 0.529 | 0.527 | 0.427 | 0.5 | 0.507 |
| Peach | 0.785 | 0.845 | 0.887 | 0.389 | 0.333 | 0.347 | 0.354 | 0.308 | 0.347 |
| Pineapple | 0.945 | 0.972 | 0.993 | 0.547 | 0.806 | 0.791 | 0.609 | 0.778 | 0.756 |
| Pomegranate | 0.82 | 0.716 | 0.802 | 0.491 | 0.471 | 0.512 | 0.599 | 0.528 | 0.544 |
| Strawberry | 0.79 | 0.838 | 0.919 | 0.421 | 0.757 | 0.466 | 0.478 | 0.791 | 0.606 |

Table 5. Fruit dataset training performance for YOLONAS.

| Model | Precision | Recall | F1 | mAP | loss_cls | loss_iou | loss_dfl | loss |
|---|---|---|---|---|---|---|---|---|
| YOLONAS_s | 0.032 | 0.112 | 0.063 | 0.809 | 0.929 | 0.112 | 1.07 | 1.748 |
| YOLONAS_m | 0.04 | 0.98 | 0.078 | 0.78 | 0.932 | 0.112 | 1.074 | 1.751 |
| YOLONAS_l | 0.034 | 0.987 | 0.0661 | 0.8326 | 0.899 | 0.116 | 1.093 | 1.736 |

Table 5 presents the training performance of the YOLONAS model using the fruit dataset. The model YOLONAS_l demonstrates the highest mean average precision (mAP) in training, achieving a value of 83.26%. Additionally, it exhibits the lowest loss value of 1.736. Following this, the model YOLONAS_s achieves a mAP of 80.9% and a loss value of 1.748.

Localization loss and classification loss are combined to form the YOLO loss function. This loss function's goal is to quantify the dissimilarity between the ground-truth annotations and the predicted bounding boxes and class probabilities. The training process of YOLO consists of the following components: The localization loss is a measure of how well the anticipated bounding box locations match the actual ones. Metrics like mean squared error (MSE) and mean absolute error (MAE) are frequently used to depict the discrepancy between the predicted box coordinates and the actual box coordinates (center coordinates, width, and height). A model's ability to forecast a confidence score for a set of bounding boxes is evaluated using a metric called "confidence loss." The IoU is calculated by comparing the projected box with the actual box. The loss may correspond to the binary cross-entropy between the confidence in the prediction and the indication of the object's presence or absence. For each bounding box, the classification loss measures how accurately the model can predict the class probabilities. This is calculated by comparing the predicted class probabilities to the actual class labels, which are presented as one-hot vectors, to get the cross-entropy loss.

The overall YOLO loss is a weighted sum of these individual loss components. The exact formulation can be different between YOLO versions, and some versions might include additional terms or regularization components. Yolo loss function based on Equation (7) [59].

$$\lambda_{coord} \sum_{i=0}^{s^2} \sum_{j=0}^{B} \mathbb{1}_{ij}^{obj}[(x_i - \hat{x}_i)^2 + (y - \hat{y}_i)^2] + \lambda_{coord} \sum_{i=0}^{s^2} \sum_{j=0}^{B} \mathbb{1}_{ij}^{obj}\left[\left(\sqrt{w_i} - \sqrt{\hat{w}_i}\right)^2 + \left(\sqrt{h_i} - \sqrt{\hat{h}_i}\right)^2\right] + \sum_{i=0}^{s^2} \sum_{j=0}^{B} \mathbb{1}_{ij}^{obj}(C_i - \hat{C}_i)^2 + \lambda_{noobj} \sum_{i=0}^{s^2} \sum_{j=0}^{B} \mathbb{1}_{ij}^{noobj}(C_i - \hat{C}_i)^2 + \sum_{i=0}^{s^2} \mathbb{1}_{i}^{obj} \sum_{c \in classes}(p_i(c) - \hat{p}_i(c))^2 \tag{7}$$

where double-struck where $\mathbb{1}_{ij}^{obj}$ indicates if the object appears in cell i, and $\mathbb{1}_{ij}^{obj}$ denotes that the $j^{th}$ bounding box predictor in cell i is responsible for the prediction. Next, $(\hat{x}, \hat{y}, \hat{w}, \hat{h}, \hat{c}, \hat{p})$ implemented to express the anticipated bounding box's center coordinates, width, height, confidence, and category probability. This experiment employed the $\lambda coord$ to 0.5, demonstrating that the width and height errors are less useful in the computation. To mitigate the effect of numerous vacant grids on the loss value, $\lambda noobj = 0.5$ is utilized.

In object detection, there can be multiple classes of objects to detect. mAP calculates the AP for each class and then computes the meaning of these AP values. This provides an overall assessment of the model's performance across all classes, accounting for the varying difficulty levels of detecting different objects. mAP is described in Equation (8).

$$mAP = \int_0^1 p(0) do \tag{8}$$

The variable p(o) represents the precision of the object detection. The Intersection over Union (IoU) metric quantifies the degree of overlap between the bounding boxes of the prediction (pred) and the ground-truth (gt), as expressed in Equation (9). Precision and recall are represented by [60] in Equation (10)-(11).

$$IoU = \frac{Area_{pred} \cap Area_{gt}}{Area_{pred} \cup Area_{gt}} \tag{9}$$

$$Precision = \frac{TP}{TP+FP} = TP/N \tag{10}$$

$$Recall = \frac{TP}{TP+FN} \tag{11}$$

where TP represents true positives, FP represents false positives, FN represents false negatives, and N represents the total number of objects recovered include the true positives and false positives. Another evaluation index, F1 [61] is shown in Equation (12).

$$F1 = \frac{2 \ x \ Precision \ x \ Recall}{Precision + Recall} \tag{12}$$

## 4. Discussion

The testing accuracy for all classes of the Fruit datasets with YOLOv8 can be found in Table 6. According to the obtained test results, it was observed that YOLOv8m achieved the highest mean average precision (mAP) of 99% for all classes, followed by the YOLOv8s with 98.9% mAP, YOLOv8n with 97.5% mAP. YOLOv5n only achieves 67.4% mAP in testing stage as we can see in Table 7.

Table 8 exhibits the testing result of YOLONAS with the fruit dataset. YOLONAS_l achieved the optimum mAP of 99.98% during the testing phase and got the minimum loss value of 0.629.

Table 6. Testing the models' accuracy with Fruit Dataset with YOLOv8.

| Class | YOLOv8n | | | YOLOv8s | | | YOLOv8m | | |
|---|---|---|---|---|---|---|---|---|---|
| | P | R | mAP | P | R | mAP | P | R | mAP |
| All | 0.957 | 0.935 | 0.975 | 0.976 | 0.972 | 0.989 | 0.975 | 0.974 | **0.99** |
| Apple B | 0.955 | 0.939 | 0.965 | 0.993 | 0.955 | 0.993 | 0.988 | 0.955 | 0.991 |
| Apple GI | 0.976 | 0.942 | 0.97 | 0.99 | 0.965 | 0.992 | 0.986 | 0.971 | 0.992 |
| Apricot | 0.927 | 0.937 | 0.977 | 0.97 | 0.985 | 0.994 | 0.971 | 0.998 | 0.994 |
| Avocado | 0.974 | 0.957 | 0.991 | 0.975 | 0.996 | 0.994 | 0.973 | 0.99 | 0.994 |
| Banana | 0.985 | 0.97 | 0.976 | 0.978 | 0.976 | 0.975 | 0.976 | 0.976 | 0.981 |
| Cherry | 0.967 | 0.972 | 0.979 | 0.981 | 0.989 | 0.985 | 0.981 | 0.989 | 0.982 |
| Grape | 0.984 | 0.981 | 0.98 | 0.952 | 0.981 | 0.974 | 0.977 | 0.981 | 0.985 |
| Kiwi | 0.868 | 0.95 | 0.96 | 0.945 | 0.984 | 0.99 | 0.938 | 0.986 | 0.99 |
| Orange | 0.963 | 0.906 | 0.973 | 0.976 | 0.946 | 0.99 | 0.975 | 0.946 | 0.991 |
| Mango | 0.947 | 0.808 | 0.939 | 0.979 | 0.905 | 0.984 | 0.947 | 0.924 | 0.985 |
| Papaya | 0.957 | 0.903 | 0.979 | 0.971 | 0.967 | 0.993 | 0.966 | 0.964 | 0.992 |
| Peach | 0.983 | 0.1 | 0.994 | 0.99 | 1 | 0.995 | 0.989 | 1 | 0.995 |
| Pineaple | 0.976 | 0.941 | 0.988 | 1 | 0.976 | 0.995 | 0.99 | 0.977 | 0.994 |
| Pomegrana | 0.922 | 0.864 | 0.959 | 0.972 | 0.979 | 0.993 | 0.974 | 0.98 | 0.994 |
| Strawberry | 0,97 | 0,948 | 0,988 | 0.974 | 0.974 | 0.994 | 0.987 | 0.974 | 0.994 |

Table 7. Testing the models' accuracy with Fruit Dataset with YOLOv5 and YOLOv7

| Class | YOLOv5n | | | YOLOv7 | | | YOLOv7x | | |
|---|---|---|---|---|---|---|---|---|---|
| | P | R | mAP | P | R | mAP | P | R | mAP |
| All | 0.618 | 0.697 | **0.674** | 0.541 | 0.555 | 0.562 | 0.464 | 0.481 | 0.468 |
| Apple B | 0.787 | 0.687 | 0.785 | 0.461 | 0.457 | 0.477 | 0.525 | 0.231 | 0.398 |
| Apple GI | 0.79 | 0.509 | 0.777 | 0.621 | 0.703 | 0.701 | 0.546 | 0.784 | 0.698 |
| Apricot | 0.472 | 0.469 | 0.378 | 0.497 | 0.346 | 0.396 | 0.318 | 0.425 | 0.344 |
| Avocado | 0.552 | 0.705 | 0.579 | 0.648 | 0.386 | 0.478 | 0.47 | 0.273 | 0.358 |
| Banana | 0.689 | 0.846 | 0.761 | 0.754 | 0.471 | 0.682 | 0.543 | 0.212 | 0.351 |
| Cherry | 0.517 | 0.583 | 0.553 | 0.56 | 0.647 | 0.645 | 0.45 | 0.404 | 0.408 |
| Grape | 0.72 | 0.87 | 0.886 | 0.617 | 0.696 | 0.749 | 0.484 | 0.609 | 0.6 |
| Kiwi | 0.763 | 0.533 | 0.676 | 0.698 | 0.578 | 0.655 | 0.649 | 0.556 | 0.59 |
| Orange | 0.48 | 0.614 | 0.481 | 0.431 | 0.409 | 0.34 | 0.427 | 0.205 | 0.231 |
| Mango | 0.525 | 0.475 | 0.567 | 0.552 | 0.732 | 0.695 | 0.419 | 0.786 | 0.572 |
| Papaya | 0.615 | 0.912 | 0.882 | 0.447 | 0.529 | 0.52 | 0.402 | 0.382 | 0.368 |
| Peach | 0.605 | 0.628 | 0.608 | 0.394 | 0.333 | 0.346 | 0.414 | 0.282 | 0.353 |
| Pineaple | 0.691 | 0.917 | 0.888 | 0.531 | 0.806 | 0.784 | 0.459 | 0.833 | 0.814 |
| Pomegrana | 0.506 | 0.824 | 0.707 | 0.493 | 0.471 | 0.506 | 0.503 | 0.431 | 0.426 |
| Strawberry | 0.563 | 0.892 | 0.588 | 0.416 | 0.757 | 0.463 | 0.345 | 0.811 | 0.512 |

Table 8. Testing the models' accuracy with Fruit Dataset with YOLONAS.

| Model | Precision | Recall | F1 | mAP | loss_cls | loss_iou | loss_dfl | loss |
|---|---|---|---|---|---|---|---|---|
| YOLONAS_s | 0.726 | 1 | 0.838 | 0.996 | 0.341 | 0.011 | 0.5285 | 0.633 |
| YOLONAS_m | 0.584 | 1 | 0.732 | 0.997 | 0.344 | 0.0104 | 0.529 | 0.635 |
| YOLONAS_l | 0.602 | 1 | 0.746 | **0.998** | 0.334 | 0.011 | 0.533 | 0.629 |

The outcomes of the fruit dataset recognition using YOLONAS are depicted in Figure 6. The model we have proposed has a high level of accuracy in detecting items inside the image. YOLONAS could differentiate between 15 distinct categories, for examples in Figure 6, the detection of YOLONAS achieved accuracy rates of 54% to 70% for Grape, 60% to 92% for Apple, 62% to 91% for Orange, and 69% to 94% for Mango.

The potential of NAS to automate the architecture design process minimizes the amount of labor necessary to manually develop the neural network structure. This is one of the benefits of neural architecture search. This is the primary benefit it offers. This automation speeds up the development process significantly, and it also enables academics to focus on other aspects of the problem, which is a win-win situation. (2) Superior high-level performance The NAS has demonstrated that it can discover unique, high-performance architectures that are superior to neural networks developed by humans in a few different contexts and applications. The field of computer vision, as well as natural language processing and picture identification, has benefited significantly from its development. (3) The capacity to scale for complex jobs as jobs get more complex and data dense, the manual design of the architecture becomes impractical. The NAS provides a fruitful way for investigating the possibilities of a diverse range of architectures that are suitable for hard and extensive endeavors.

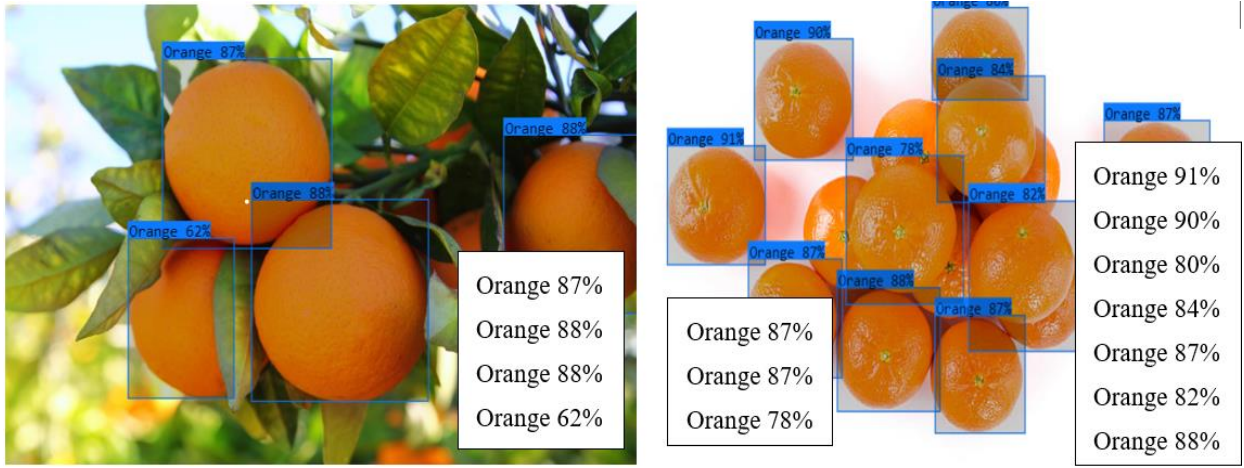

(a) Orange

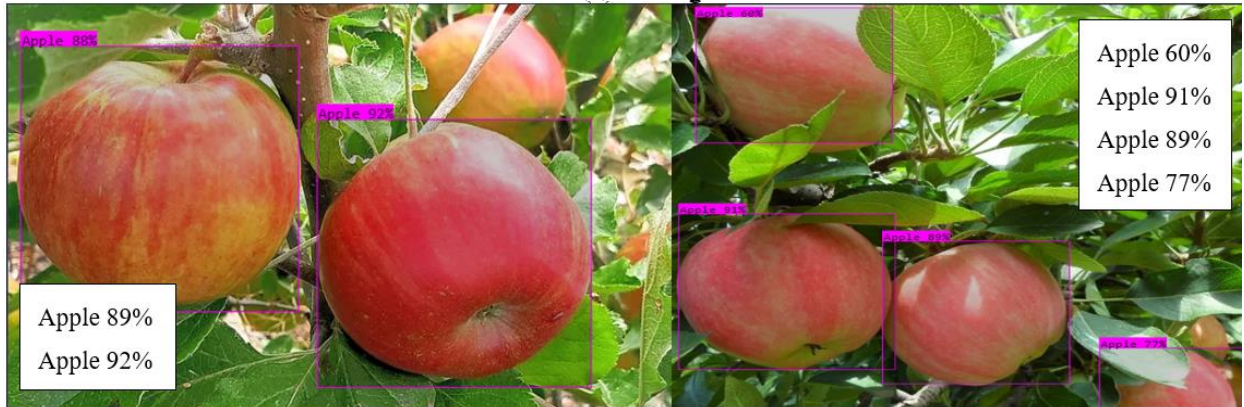

(b) Apple

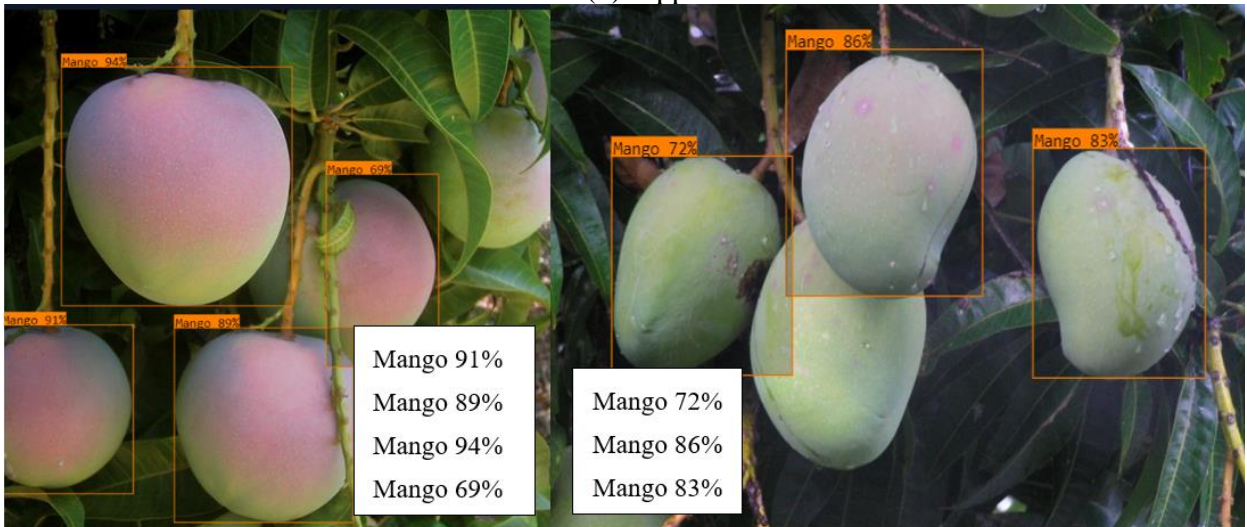

(c) Mango

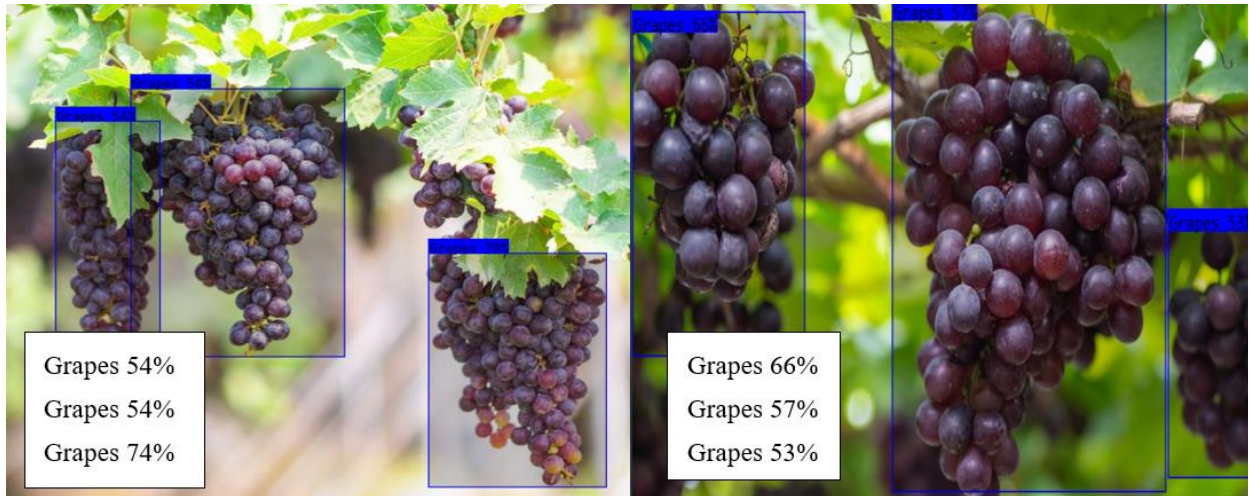
(d) Grape

Figure 6. Fruit Recognition Result using YOLONAS_l class (a) Orange, (b) Apple, (c) Mango, and (d) Grape.

Table 9. Previous Research Comparison with Fruit Dataset.

| Reference | Year | Method | Result AP (%) |
|---|---|---|---|
| He, K. et al., [62] | 2016 | PCA+ FSCAB | 87.75 |
| Zhang, Y. et al., [63] | 2014 | WE+BBO | 87.17 |
| Wang, S. et al., [64] | 2015 | FRFE+BPNN | 87.52 |
| Zhang, Y. and Wu, L., [65] | 2012 | Max-Wins-Voting (MWV-SVM) | 88.2 |
| Lu, Z. et al., [66] | 2017 | 13 Layer CNN | 92.38 |
| Xue, G. et al., [67] | 2023 | CAE+ADN | 93.78 |
| Proposed Method | 2024 | YOLONAS_l | 99.98 |

Table 9 has a description of the comparison to the previous study. With an accuracy of 99.98%, our proposed YOLONAS approach with 100 epochs exceeds earlier models that were tested on the Fruit dataset. We were able to boost the overall performance of a recent study on fruit detection in this research. He, K. et al., [62] proposed the PCA and FSCAB with only 87.75% mAP. Another researcher Zhang, Y. et al., [63] implemented WE+BBO and achieves 87.17 % mAP. Furthermore, Wang, S. et al., [64], implement FRFE+BPNN and achieves 87.52% mAP. Next, Zhang, Y. and Wu, L., [65] proposed Max-Wins-Voting (MWV-SVM) with 88.2% mAP. Lu, Z. et al., [66] employed 13 Layer CNN and achieved 92.38%, Xue, G. et al., [67] used CAE+AND and achieved 93.78%. Our proposed method YOLONAS_l with 100 epoch achieves the highest performance with 99.98% mAP.

YOLO-NAS is not only compatible with quantization but also allows the deployment of TensorRT, which ensures that it is fully compatible with production use. This breakthrough in object detection has the potential to stimulate new research and change the area. It will ultimately make it possible for machines to see and interact with the world more intelligently and autonomously. The YOLO-NAS and YOLOv8 object identification models are both highly effective and efficient in their respective applications. YOLO-NAS, on the other hand, is superior to YOLOv8 in a few important aspects, such as the identification of small objects, the precision of localization, the post-training quantization, and real-time edge-device applications. YOLO-NAS is the model that emerges as the clear victor when we consider our requirements for a cutting-edge object identification tool to include higher accuracy, faster processing, and greater efficiency.

**5. Conclusions**

In this article, the findings of research into CNN-based object identification algorithms, specifically YOLONAS, YOLOv8, YOLOv7, and YOLOv5, are presented along with a discussion of those findings. YOLO Labs is responsible for the development of these algorithms. According to the results of our examinations, the YOLONAS provides a degree of precision that is far superior to that of anything else that is currently on the market. In the context of this study, we present a unique model for the detection of fruit. In addition, the YOLONAS method with 100 epochs, which we have proposed, outperforms earlier models that have been applied to the Fruit dataset in terms of mAP. With an accuracy rate that is, on average, 99.98%. We have provided evidence to show that the YOLONAS model scheme that we have proposed is a reliable model for identifying Fruit. The YOLO-NAS model stands out as the prevailing champion when evaluating our criteria for an advanced object recognition tool, including enhanced precision, expedited processing, and improved efficiency. In addition, forthcoming research will study whether it is possible to use deep learning models for fruit identification with a wide variety of variants. In addition, we are going to look at the possibility of using explainable artificial intelligence (XAI) in fruit identification.

**Data Availability Statement:** Fruit 360 Dataset (https://www.kaggle.com/datasets/moltean/fruits (accessed on 12 May 2023))
**Acknowledgments:**
The author would like to thank all colleagues from Deakin University, Satya Wacana Christian University, Indonesia, and all involved in this research.

**Conflicts of Interest:** The authors declare no conflicts of interest.